\documentclass[sigconf]{acmart}
\usepackage{hyperref}
\usepackage{multirow}

\AtBeginDocument{%
  \providecommand\BibTeX{{%
    \normalfont B\kern-0.5em{\scshape i\kern-0.25em b}\kern-0.8em\TeX}}}

\copyrightyear{2022}
\acmYear{2022}
\setcopyright{acmcopyright}\acmConference[MM '22]{Proceedings of the 30th ACM
International Conference on Multimedia}{October 10--14, 2022}{Lisboa, Portugal}
\acmBooktitle{Proceedings of the 30th ACM International Conference on Multimedia
(MM '22), October 10--14, 2022, Lisboa, Portugal}
\acmPrice{15.00}
\acmDOI{10.1145/3503161.3547761}
\acmISBN{978-1-4503-9203-7/22/10}

\author{Weidong Chen$^{1, \dagger}$, Dexiang Hong$^{1, \dagger}$, Yuankai Qi$^{2}$, Zhenjun Han$^{1}$, Shuhui Wang$^{3, 4}$, Laiyun Qing$^{1}$}
\author{Qingming Huang$^{1, 3, 4}$, Guorong Li$^{1, *}$}
\thanks{$^\dagger$Equal contribution.}
\thanks{$^\ast$Corresponding author.}
\affiliation{%
	\institution{$^1$University of Chinese Academy of Science, Beijing, China}
    \institution{$^2$Australian Institute for Machine Learning, The University of Adelaide \quad $^3$Peng Cheng Laboratory, Shenzhen, China}
	\institution{$^4$Key Lab of Intelligent Information Processing, ICT, CAS, Beijing, China}
}
\email{weidong.chen@vipl.ict.ac.cn, {hongdexaing19, hanzj, lyqing, qmhuang, liguorong}@ucas.ac.cn}
\email{wangshuhui@ict.ac.cn, qykshr@gmail.com}

\begin{document}
\fancyhead{}

\title{Multi-Attention Network for Compressed Video Referring Object Segmentation}


\renewcommand{\shortauthors}{ }

\begin{abstract}
  
  Referring video object segmentation aims to segment the object referred by a given language expression.
  Existing works typically require compressed video bitstream to be decoded to RGB frames before being segmented, which increases computation and storage requirements and ultimately slows the inference down.
  This may hamper its application in real-world computing resource limited scenarios, such as autonomous cars and drones.
 To alleviate this problem, in this paper, we explore the referring object segmentation task on compressed videos, namely on the original video data flow.
 Besides the inherent difficulty of the video referring object segmentation task itself, obtaining discriminative representation from compressed video is also rather challenging.
 %
 To address this problem, we propose a multi-attention network which consists of dual-path dual-attention module and a query-based cross-modal Transformer module. Specifically, the dual-path dual-attention module is designed to extract effective representation from compressed data in three modalities, i.e., I-frame, Motion Vector and Residual. The query-based cross-modal Transformer firstly models the correlation between linguistic and visual modalities, and then the fused multi-modality features are used to guide object queries to generate a content-aware dynamic kernel and to predict final segmentation masks. Different from previous works, we propose to learn just one kernel, which thus removes the complicated post mask-matching procedure of existing methods. Extensive promising experimental results on three challenging datasets show the effectiveness of our method compared against several state-of-the-art methods which are proposed for processing RGB data. Source code is available at: \href{https://github.com/DexiangHong/MANet}{https://github.com/DexiangHong/MANet}.
\end{abstract}




\maketitle

\section{Introduction}

In recent years, online video has shown an exponential explosive growth trend. It has become more and more unrealistic to artificially watch and process such a tremendous amount of video data. 
With the further demand for computers to automatically analyze, understand, and process video content, many video understanding problems \cite{qzhb2020ckmn, srl, qzhb2020tdc} in deep learning and computer vision arise and thrive, such as video visual question answering \cite{li2020boosting, liu2020cascade, cui2021rosita, han2021greedy, han2020interpretable} and language-guided video action localization \cite{qu2020fine, cao2020strong}. Referring video object segmentation aims to selectively segment one specific object spatially and temporally in a video 
according to a language query. This task helps the computer to understand videos better in an artificially interactive way.


\begin{figure}[t]
\centering
\includegraphics[width=0.9\linewidth]{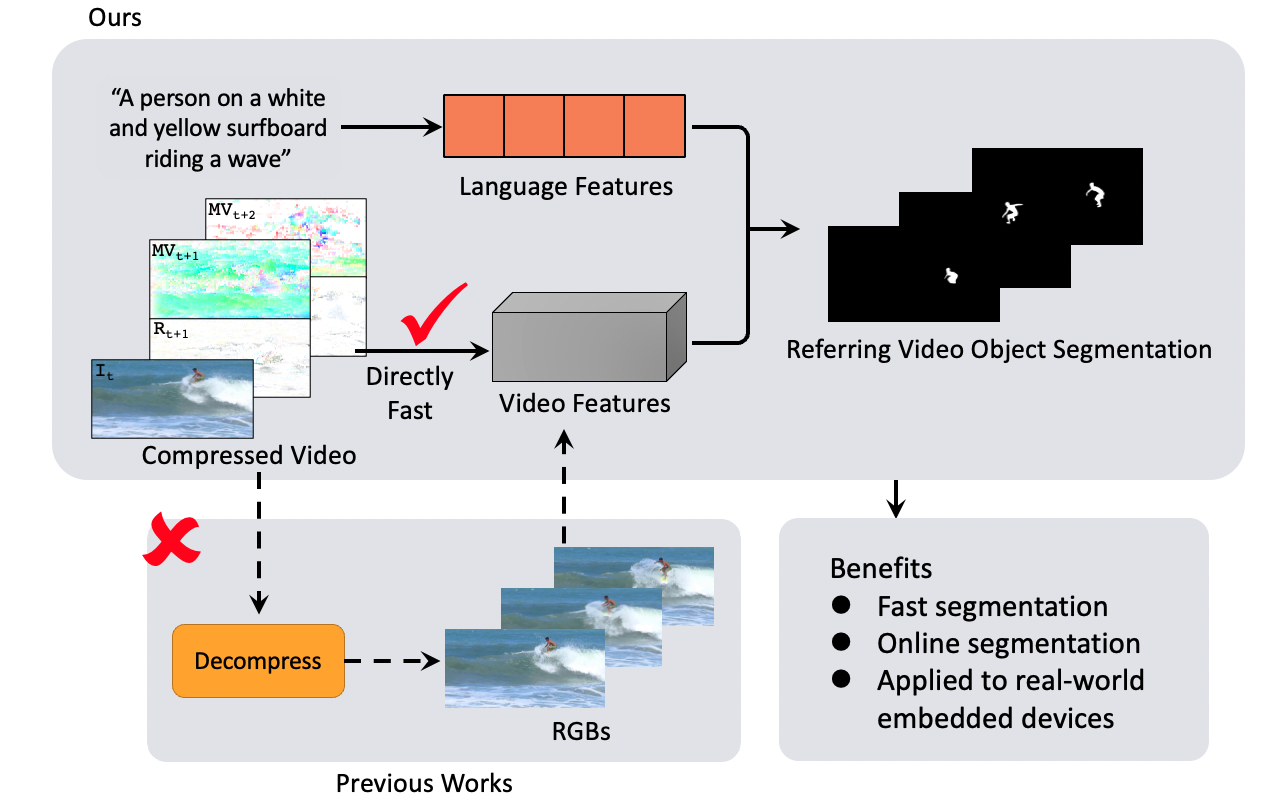}
\caption{Illustration of our motivation. The proposed method can be directly applied to the compressed video bitstreams which is able to save computational and storage resources on embedded devices.}
\vspace{-12pt}
\label{Motivation}
\end{figure}

Our motivation is illustrated in Fig.~\ref{Motivation}. Existing works typically require compressed video bitstreams to be decoded into RGB frames before being processed, which requires local storage space and increases computational cost. Meanwhile, when modeling the interaction between vision and language, most existing works \cite{gavrilyuk2018actor, chen2021cascade, wang2020context, yang2021hierarchical} leverage additional networks to estimate objects' motion, while ignoring the motion information that already exists in the compressed video bitstreams. 
Furthermore, in some real-world application scenarios, such as autonomous vehicles, online decoding video and then processing can hardly achieve real-time segmentation because decoding videos in the mobile terminal takes much time and computational cost\footnote{\url{https://developer.ridgerun.com/wiki}}. 
Thus, in this paper, we propose to accomplish referring video object segmentation directly in the compressed videos domain. Specifically, we target at MPEG-4\cite{mpeg} Part 2 compressed videos in this paper, which is composed of a set of GoPs (Group of Pictures). For each GoP, it starts with one I-frame (reference frame) and appends a series of the P-frame (predictive frames) behind. Each the P-frame consists of Motion Vector and Residual, which are used to reconstruct the RGB data of the P-frame based on I-frame \cite{li2020slow}.

Given the bitstream of the compressed video, the main problem is how to extract features from compressed video for segmentation. Different form RGB frames, three data modalities i.e. the P-frame, Motion Vector ,and Residual, in the compressed video have a strong dependency. For example, Motion Vector describes the displacement of the current frame relative to I-frames, while Residuals in the P-frames maintain the RGB differences between I-frame and its reconstructed frame calculated by Motion Vectors in the P-frames after motion compensation. Simply treat each data modality as a single data bitstream to extract features cannot extract powerful features from compressed video for segmentation. Meanwhile, Motion Vector actually has a much lower resolution than I-frames, since the values of Motion Vector within the same macro-block is the same, leading the features extracted from the P-frame are not strong enough on the spatial dimension, which further affects the performance of segmentation.

To address this problem, we propose a novel dual-path dual-attention module. 
First, the proposed module leverages two dual-attention in a parallel manner to get the I-frame attended Motion Vector features and attended the Residual features, respectively. 
We then incorporate the attended Motion Vector features and the Residual features to get powerful the P-frame features for modeling cross-modal correlation and segmentation.
Second, 
Dual attention measures the channel-wise gated attention and spatial-wise gated attention of I-frame data modality and Residuals (Motion Vectors) data modality at first, and then, dual attention integrates the I-frame features into the Residuals (Motion Vectors) features by multiplying channel gated attention and spatial attention in a cascade manner to get the attended Residual (Motion Vectors) features.  It is worth noting that dual-attention is more computational efficient than full attention with the same performance.

Moreover, due to the diversity in language descriptions and video content, it is hard to establish a consistent correspondence between videos and texts, resulting in unavoidable undesired segmentation results.
%
%
To address this problem, we propose a query-based cross-modal Transformer. Specifically, the query-based cross-modal Transformer first fully exchanges information between two modalities 
to bridge the gap between two modalities.
Finally, the multi-modality features guide object queries to generate a content-aware dynamic kernel, and then predict final masks. 
We equip the proposed Transformer with a new referring segmenting scheme, which leverages language to select one target kernel and thus leads to only one segmentation mask. This scheme dramatically reduces the number of input object queries and completely removes the complicated mask-matching procedure of the existing method and, therefore, boosts the speed.

In summary, our main contributions are summarized as follows:

$\bullet$ To the best of knowledge, we are the first to explore the video referring object segmentation task in the compressed domain. We propose a multi-attention network, which is formed by two modules, i.e., dual-path dual-attention module and query-based cross-modal Transformer, and achieves favourable results compared against several state-of-the-art methods in RGB domain.

$\bullet$ The dual-path dual-attention module is designed to learn powerful representations of compressed videos. The proposed module leverages two dual-attention in a parallel manner to get the I-frame attended Motion Vector features and the Residual features, respectively; whereas, dual-attention integrates the I-frame features into the Residuals (Motion Vectors) features by multiplying channel gated attention and spatial attention in a cascade manner.

$\bullet$ We propose a query-based cross-modal Transformer, and equip it with a new referring segmenting scheme, which leverages language to select one target kernel and thus leads to only one segmentation mask. This scheme dramatically reduces the number of input object queries and completely removes the complicated mask-matching procedure of the existing method and, therefore, boosts the speed.

\section{Related Work}

\subsection{Text-based video object localization}

Text-based video object localization is a problem that connects computer vision and natural language processing. 
Xu \textit{et al.} \cite{xu2015can, xu2016actor, yan2017weakly} propose an actor-action semantic segmentation task, and introduces the challenging large-scale A2D dataset.
To further study cross-modal video understanding, Gavrilyuk \textit{et al.} \cite{gavrilyuk2018actor} augment A2D dataset by further adding human generated description of a specific object of each video. The new task can be modeled as spatio-temporal video object localization. 
Several works \cite{gavrilyuk2018actor, wang2020context, hui2021collaborative, botach2021end} focus on dynamic convolution-based method; while the last two methods incorporate spatial context into filter generation. Several works \cite{wang2019asymmetric, liu2021cross, ye2021referring, seo2020urvos, chen2021cascade} focus on attention-based method. McIntosh \textit{et al.} \cite{mcintosh2020visual} further encode the visual and language features as capsules and integrate the visual and language information via a routing algorithm.
%
The very recent method MTTR \cite{botach2021end} leverages Transformer to do referring video object segmentation, and achieves remarkable performance. However, it requires well-prepared frame-level annotations to do instance segmentation first before segmenting referring object, and it needs a complex post-processing to select one instance sequential mask as referred one, which slows the inference down. Unlike previous works that focus on decompressed video frames, in this work, we aim to investigate compressed video referring object segmentation. It benefits us on several computing resource limited scenarios, such as online segmentation, real-time fast segmentation and run on embedded devices.

\subsection{Compressed Video Action Recognition}

A lot of works have achieved great success in compressed video action recognition, they give us a great inspiration to extract powerful features from compressed video. In the pioneering works \cite{zhang2016real, zhang2018real}, motion vectors are utilized to replace optical flow features. However, they leverage both compressed videos and raw videos simultaneously to do prediction. In order to improve the performance, some methods introduce optical flows into features from compressed video as extra information \cite{shou2019dmc, wu2018compressed}. 
Li \textit{et al.} \cite{li2020slow} propose a novel Slow-I-Fast-P (SIFP) neural network model for compressed video action recognition. It consists of the slow I pathway receiving a sparse sampling I-frame clip and the fast P pathway receiving a dense sampling pseudo optical flow clip.
Hu \textit{et al.} \cite{hu2020mv2flow} train four independent classification networks and combine them by late fusion to make the final prediction.
Unlike existing approaches that encode I-frames and the P-frames individually, Yu \textit{et al.} \cite{yu2020self} propose to jointly encode them by establishing bidirectional dynamic connections across streams.
Inspired by previous works, we jointly encode three data modalities features to learn powerful video features for segmentation in this work.

\subsection{Attention}

Recently, attention mechanisms \cite{zhuo2017deep, wang2018joint} have been widely-used in visual-language task, such as VQA \cite{xu2015show,2016stacked_attention,2016hierarchical_attention,2019inter_intra,2018BAN, lu2018co, han2020interpretable, han2021greedy} and referring expression comprehension \cite{ye2019cross, wang2018non}.
Fu \textit{et al.} \cite{fu2019dual} develop a dual attention network, which contains two parallel attention modules: one is for spatial-attention and another for channel-wise attention.
Tsai \textit{et al.} \cite{tsai2019multimodal} introduce the cross-modal attention network to model word-based spatial attention, and further designs a novel transformer based on cross-modal attention.
Ye \textit{et al.} \cite{ye2019cross} propose a cross-modal self-attention to do referring segmentation in image-level.
Wang \textit{et al.} \cite{wang2019asymmetric} design a asymmetric cross-modal attention module, which uses not only the gated self-attention to model the language's impact on vision, but also co-attention to model the visual impact on language. Inspired by previous works that leverage dual attention to model cross-modalities correlations, we propose a dual-path dual-attention module in this paper. This module integrate the I-frame features into the Residuals and Motion Vectors by cascade dual-attention, including channel attention and spatial attention. By getting I-frame features attended the Residual features and Motion Vector features in a parallel manner and then fusing, the P-frame features are becoming powerful for segmentation. 

\subsection{Multi-modal Transformer}

The encoder-decoder architecture of the Transformer can be adopted to multi-modal tasks such as captioning, question-answering, reasoning, and visual grounding. Vision-and-language pre-trained model is becoming a trend in this field, such as \cite{li2019visualbert, su2019vl, lu2019vilbert}.
Sun \textit{et al.} \cite{sun2019videobert} propose an excellent work, VideoBERT, which learns joint video-text representations with Transformer in a self-supervised manner for downstream tasks. A multi-modal Transformer is proposed in \cite{yu2019multimodal} for image captioning. Object proposals generated from multiple detectors are fed into the Transformer encoder, and the Transformer decoder learns a language model conditioned on the encoder outputs. In referring expression comprehension task, Sun \textit{et al.} \cite{suo2021proposal} propose a set of Transformers to localize referred objects in a one-stage manner. 
Li \textit{et al.} \cite{li2021referring} jointly train their proposed novel Transformer for referring expression segmentation and comprehension tasks while enabling contextualized multi-expression references. Recently, MTTR \cite{botach2021end} and MDETR \cite{kamath2021mdetr} leverage the successful encoder-decoder architecture of the Transformer to design their own methods for referring video/image segmentation. They develop their methods based on DETR \cite{carion2020end} and VisTR \cite{wang2021end}. However, both two methods require a large number of object queries, and well-prepared annotations to train instance segmentation first.

\section{Proposed Method}

\begin{figure*}[t]
\centering
\includegraphics[width=0.95\linewidth]{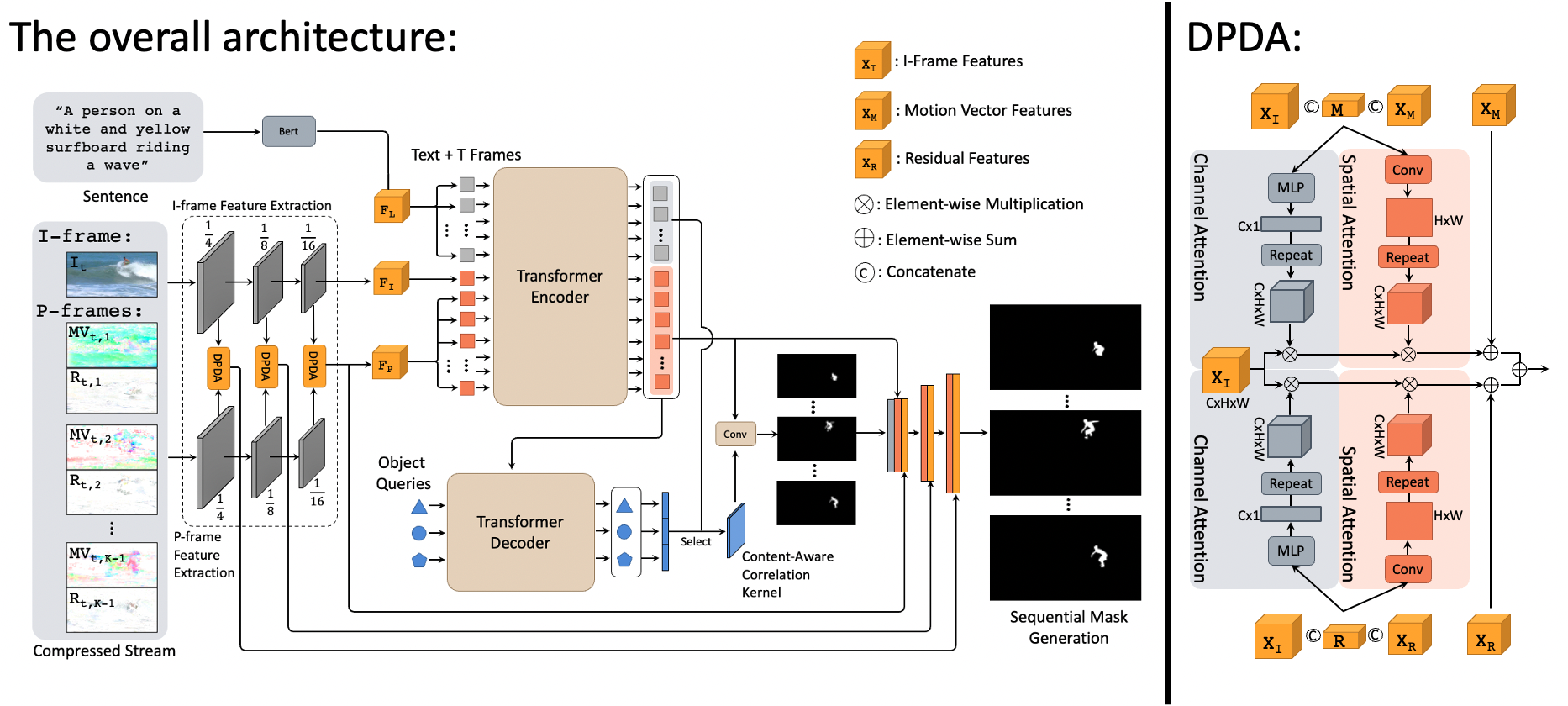}
\caption{The overall architecture of our proposed method. It contains two major modules, including dual-path dual-attention module and query-based cross-modal Transformer. Some notations are illustrated in the framework. DPDA stands for dual-path dual-attention module.}
\label{fig3_1}
\end{figure*}

As described above, we design a multi-attention network for Compressed Video Referring Object
Segmentation. Specifically, Following previous works \cite{shou2019dmc, wu2018compressed, li2020slow, hu2020mv2flow, yu2020self}, we mainly process the MPEG-4 Part 2 compressed videos. 
Formally, our input is $T$ GoPs, $G_1$, · · · , $G_T$, where each $G_t$ contains one I-frame $I_t \in \mathbb{R}^{H\times W\times 3}$ followed by $K$ pairs of motion vectors $M_{t, k} \in \mathbb{R}^{H\times W\times 2}$ and residuals $R_{t, k} \in \mathbb{R}^{H\times W\times 3}$, $k \in [1, K]$. For efficiency and simplicity, we assume an identical GoP size $K$ for all $t \in [1, T]$. The target is to segment video objects in each ($T\times K$) frames based on the language descriptions.
As shown in Fig. \ref{fig3_1}, the proposed method is formed by two modules, i.e., a dual-path dual-attention module and a query-based cross-modal Transformer. The dual-path dual-attention aims to fast incorporate three data modalities to extract powerful video features from compressed video. Meanwhile, powerful video features from the compressed video and the language features are fed into the query-based cross-modal Transformer. The Transformer encoder wholly exchanges cross-modal information at first, and then multi-modal sequences guide object queries in the Transformer decoder to learn content-aware kernels and predict masks. The proposed method equips the proposed Transformer with a new referring segmenting scheme that sorts kernels and selects a best-matched kernel before generating masks to avoid the complicated mask-matching procedure of the existing method and therefore boosts the speed.

\textbf{Visual Encoder.} 
Considering the powerful performance of Video-Swin-T \cite{liu2021swin, liu2021video}, we adopt it to extract three scale visual features, denoted as $X_{I_s}$, $X_{I_m}$, $X_{I_l}$, respectively.
of each reference frame (I-frame) among all I-frames in the video clip input with no temporal downsampling. For Motion Vectors and Residuals in each the P-frame, following previous works \cite{shou2019dmc, wu2018compressed}, we use ResNet-18 \cite{he2016deep} to extract three scale visual features, denoted as $X_{M_s}$, $X_{M_m}$, $X_{M_l}$, and $X_{R_s}$, $X_{R_m}$, $X_{R_l}$, respectively. 

\textbf{Linguistic Encoder.} Given the language description with $N$ words, we use off-the-shelf linguistic embedding model, BERT \cite{devlin2018bert}, to extract the text feature $F_L \in \mathbb{R}^{L\times C_L}$. Each row represents a word features denoted by $F_L^i \in \mathbb{R}^{C_L}$, where $i={1, 2, ..., N}$.

\subsection{Dual-path Dual-attention Module}



Because of the lightweight the P-frame feature extractor ResNet-18, the feature map of the P-frame is not strong enough for segmentation. Meanwhile, since the Residuals in the P-frames maintain the RGB differences between I-frame and its reconstructed frames calculated by Motion Vectors in the P-frames after motion compensation, both Motion Vectors and Residuals are important for extracting powerful the P-frame features. Thus, we design a dual-path dual-attention module to encode powerful features from compressed video for segmentation.



\noindent{\textbf{$\bullet$ Dual-Attention in MV->I Stream}}

Take one GoP and the last scale as an example. Firstly, for k-th the P-frame, we project the channel dimension of I-frame feature $X_{I_s}$ into the same channel dimension of Motion Vector features $X_{M_s}^k$. 
Meanwhile, to avoid losing part of Motion Vector information during feature extraction, we concatenate the spatial down-sampling Motion Vector $M^k$, Motion Vector features $X_{M_s}^k$ and I-frame features $X_{I_s}$ together first.
Then, we leverage ${\rm Conv_D}$ to fully exchange and fuse two data modalities, I-frames features and Residuals. ${\rm Conv_D}$ is a dense-connected four convolutional layers, details are shown in \ref{ImpleDets}. 
Thus, the formulation is:
\begin{equation}
Y_{IM_s}^k = {\rm Conv_{D}}([X_{I_s}; X_{M_s}^k; M^k])
\label{E1}
\end{equation}
where $Y_{IM_s}^k$ is the fused features of I-frame and Motion Vectors, $[; ]$ is the concatenation operation.

Then, we leverage a fully connected layer to project the fused features to obtain the channel attention between I-frame and Motion Vectors:
\begin{equation}
\overline{Att}_{{\rm cha}, IM_s}^k = \sigma_s(W_C \cdot Y_{IM_s}^k) 
\label{equation3_2_2}
\end{equation}
where $W_C$ is the learnable coefficient, and $\sigma_s$ is the softmax activation. $\overline{Att}_{{\rm cha}, IM_s}^t$ represents gated attention on the channel dimension with each element measuring the impact of I-frame channel on Motion Vector channel.
We duplicate $\overline{Att}_{{\rm cha}, IM_s}^k$ into the same size of I-frame features $X_{I_s}$, and element-wise multiple attention map with the $X_{I_s}$ to get the attend Motion Vector features $X_{{\rm cha}, IM_s}^k$ in channel dimension:
\begin{equation}
X_{{\rm cha}, IM_s}^k = X_{I_s} \odot Att_{{\rm cha}, IM_s}^k
\end{equation}
where $\odot$ represents element-wise multiplication.

Meanwhile, in spatial attention, we firstly leverage a simple convolutional layer to get the attention map $\overline{Att}_{{\rm spa}, IM}^k$ on spatial dimension between I-frame and Motion Vector. Each element in the attention map measuring the impact of I-frame spatial positions on Residual spatial position:
\begin{equation}
\overline{Att}_{{\rm spa}, IM_s}^k = \sigma_s(w_{{\rm spa}} \otimes Y_{IM_s}^k) 
\label{equation3_2_3}
\end{equation}
where $w_{spa}$ is the convolutional kernel, $\otimes$ means convolution operation. 
Then, duplicate the $\overline{Att}_{{\rm spa}, IM}^t$ into the same dimension of $X_I$, and element-wise multiple with the channel attended features to get attended Motion Vector features of I-frame:
\begin{equation}
X_{{\rm spa}, IM_s}^k = X_{{\rm cha},IM_s}^k \odot Att_{{\rm cha}, IM_s}^k
\end{equation}

Thus, the total formulation is:
\begin{equation}
F_{M_s}^k = Att_{{\rm spa}, IM_s}^k \odot (Att_{{\rm cha}, IM_s}^k \odot X_{I_s}) + X_{M_s}^k
\label{equation3_2_4}
\end{equation}
where $F_{M_s}^k$ is the final attended visual features in Motion Vector path. 

\noindent{\textbf{$\bullet$ Dual-Attention in R->I Stream}}

Residuals 
maintain the RGB differences between 
I-frame and its reconstructed frame calculated by Motion Vectors 
after motion compensation. Therefore, we leverage the same dual-attention operation as Motion Vectors branch to integrate the Residual features into the P-frames features extraction. Similarly, the general formulation in Residual path is:
\begin{equation}
F_{R_s}^k = Att_{{\rm spa}, IR_s}^k \odot (Att_{{\rm cha}, IR_s}^k \odot X_{I_s}) + X_{R_s}^k
\label{equation3_2_4}
\end{equation}
where $F_{R_s}^k$ represents the final attended visual features in Residual path.

\noindent{\textbf{$\bullet$ Dual-path Dual-Attention Fusion}}

After getting I-frame features attended Motion Vector features and the Residual features, both two are important for powerful the P-frame features extraction. We leverage a simple summation operation to fuse Motion Vector and Residual to get the P-frame features $F^k$: 
\begin{equation}
F_s^k = F_{M_s}^k+F_{R_s}^k
\label{equation3_2_5}
\end{equation}
where $F_s^k$ is the augmented features of the $k$-th the P-frame.
Meanwhile, our proposed dual-path dual-attention calculation lies in shadow convolution layers and element-wise multiplication, which is computational efficient.

The above illustration is an example of one GoP. Thus, generally, $F_s^k$ is actually $F_s^{t,k}$. Thus, the output after passing dual-path dual-attention is $F_s = [F_s^{1,1}, F_s^{1,2}, ..., F_s^{1,K}, F_s^{2,1}, F_s^{2,2}, ..., F_s^{T,K}]$. Then, we utilize a 4-layer 3D-CNN to downsample the temporal dimension to get clip features $V\in \mathbb{R}^{\frac{H}{16}\times \frac{W}{16}\times C_V}$, and then, we concatenate clip features in the channel dimension to each frame features to get the overall video features $F_V \in \mathbb{R}^{(T\times K)\times \frac{H}{16}\times \frac{W}{16 }\times (2C_V)}$. Meanwhile, there are another two scales dual-path dual-attention outputs $F_m$ and $F_l$, which are used for mask generation.

\subsection{Query-based Cross-modal Transformer}

The proposed query-based cross-modal Transformer firstly leverages the Transformer encoder to fully exchange two modalities and reduce modalities’ gap, and generates language-guided visual features $F_{L\rightarrow V}$ and vision-guided linguistic features $F_{V\rightarrow L}$. 
Both focus on referring video objects. Then, the cross-modal features along with the object queries are fed into to the Transformer decoder, guiding 
the object queries focusing on learning the content of referring object from multi-modal perspectives. 


\noindent{\textbf{$\bullet$ Transformer Encoder}}




The proposed query-based cross-modal Transformer employs the feature map of a video clip $F_V \in \mathbb{R}^{(T\times K)\times \frac{H}{16}\times \frac{W}{16}\times 2C_V}$ and language embedding vector $F_L \in \mathbb{R}^{N\times C_L}$ as inputs.
These visual and linguistic features are linearly projected to a shared dimension $D$ at first. Then, we add a fixed 2D positional encoding to the feature map of each frame and the features of each word, following \cite{botach2021end}. The features of each frame are then flattened and concatenated with text embeddings respectively, resulting in multi-modal sequence whose size is $((T\times K)\times \frac{H}{16}\times \frac{W}{16}+N)\times D$.

\noindent{\textbf{$\bullet$ Transformer Decoder}}

The multi-modal sequence along with $N_q$ object queries are then fed into the Transformer decoder. The multi-modal sequence highlights the referring video objects in both visual and linguistic aspects, and the multi-modal sequence points out the learning keypoint of the object queries. Unlike previous work \cite{botach2021end}, our method sort and select the dynamic kernels before predicting the sequential masks, which greatly reduces the number of object queries, avoids post-processing to link each video object in each frame, so as to save a lot computational cost. 

\subsection{Mask Generation}

\begin{table*}[t]
\centering
\resizebox{0.8\linewidth}{!}{
    \begin{tabular}{cccccccccc}
    \hline
    \hline
    \multirow{2}{*}{Method} &\multirow{2}{*}{Venue} &\multirow{2}{*}{P@0.5} &\multirow{2}{*}{P@0.6} &\multirow{2}{*}{P@0.7} &\multirow{2}{*}{P@0.8} &\multirow{2}{*}{P@0.9}  &\multirow{2}{*}{mAP}  &\multicolumn{2}{c}{IoU} \\
    \cline{9-10} & & & & & & &  &Overall &Mean\\
    \hline
    A2DS \cite{gavrilyuk2018actor}    &CVPR18          &50.0 &37.6 &23.1 &9.4  &0.4 &21.5 &55.1 &42.6\\
    ACGA \cite{wang2019asymmetric}         &ICCV19          &55.7 &45.9 &31.9 &16.0 &2.0 &27.4 &60.1 &49.0\\
    CMDy \cite{wang2020context}            &AAAI20          &60.7 &52.5 &40.5 &23.5 &4.5 &33.3 &62.3 &53.1\\
    VT-Capsule \cite{mcintosh2020visual}     &CVPR20          &52.6 &45.0 &34.5 &20.7 &3.6 &30.3 &56.8 &46.0\\
    PolarRPE \cite{ning2020polar}              &IJCAI20         &63.4 &57.9 &48.3 &32.2 &8.3 &38.8 &66.1 &52.9\\
    CSTM \cite{hui2021collaborative}        &CVPR21          &65.4 &58.9 &49.7 &33.3 &9.1 &39.9 &66.2 &56.1\\
    CMPC-V \cite{liu2021cross}       &T-PAMI21  &65.5 &59.2 &50.6 &34.2 &9.8 &40.4 &65.3 &57.3\\
    CMSA+CFSA \cite{ye2021referring} &T-PAMI21  &48.7 &43.1 &35.8 &23.1 &5.2 &-    &61.8 &43.2\\
    CCMA \cite{chen2021cascade} &ACMMM21  &65.3 &64.5 &\textbf{61.1} &\textbf{49.1} &\textbf{17.4} &\textbf{48.0} &63.2 &55.5\\
    MTTR \cite{botach2021end} &CVPR22        &\textbf{72.1} &\textbf{68.4} &\textbf{60.7} &\textbf{45.6} &\textbf{16.4} &44.7   &\textbf{70.2} &\textbf{61.8}\\
    \hline
    Ours &-  &\textbf{73.4} &\textbf{68.2} &57.9 &38.9 &13.2 &\textbf{47.1} &\textbf{72.6} &\textbf{63.2}\\
    \hline
    \hline
\end{tabular}}
\caption{Comparison with the state-of-the-art methods on A2D Sentences. Our proposed model outperforms the state-of-the-art methods for nearly all metrics. Top-2 results are highlighted in bold.}
\label{Table1}
\vspace{-5pt}
\end{table*}

\begin{table*}[t]
\centering
\resizebox{0.8\linewidth}{!}{
    \begin{tabular}{cccccccccc}
    \hline
    \hline
    \multirow{2}{*}{Method} &\multirow{2}{*}{Venue} &\multirow{2}{*}{P@0.5} &\multirow{2}{*}{P@0.6} &\multirow{2}{*}{P@0.7} &\multirow{2}{*}{P@0.8} &\multirow{2}{*}{P@0.9}  &\multirow{2}{*}{mAP}  &\multicolumn{2}{c}{IoU} \\
    \cline{9-10} & & & & & & &  &Overall &Mean\\
    \hline
    A2DS \cite{gavrilyuk2018actor}    &CVPR18          &69.9 &46.0 &17.3 &1.4 &0.0 &23.3 &54.1 &54.2\\
    ACGA \cite{wang2019asymmetric}         &ICCV19          &75.6 &56.4 &28.7 &3.4 &0.0 &28.9 &57.6 &58.4\\
    CMDy \cite{wang2020context}            &AAAI20          &74.2 &58.7 &31.6 &4.7 &0.0 &30.1 &55.4 &57.6\\
    VT-Capsule \cite{mcintosh2020visual}     &CVPR20          &67.7 &51.3 &28.3 &5.1 &0.0 &26.1 &53.5 &55.0\\
    PolarRPE \cite{ning2020polar}              &IJCAI20         &69.1 &57.2 &31.9 &6.0 &\textbf{0.1} &29.4 &- &- \\
    CSTM \cite{hui2021collaborative}        &CVPR21          &78.3 &63.9 &37.8 &7.6 &0.0 &33.5 &59.8 &60.4\\
    CMPC-V \cite{liu2021cross}                &T-PAMI21        &81.3 &65.7 &37.1 &7.0 &0.0 &34.2 &61.6 &61.7\\
    CMSA+CFSA \cite{ye2021referring}              &T-PAMI21        &76.4 &62.5 &38.9 &9.0 &\textbf{0.1} &-    &62.8 &58.1\\
    CCMA \cite{chen2021cascade} &ACMMM21  &86.9 &80.5 &\textbf{61.4} &\textbf{16.7} &0.0 &\textbf{44.7} &\textbf{70.2} &64.9\\
    MTTR \cite{botach2021end} &CVPR22        &\textbf{91.0} &\textbf{81.5} &\textbf{57.0} &\textbf{14.4} &\textbf{0.1} &36.6   &67.4 &\textbf{67.9}\\
    \hline
    Ours              &-  &\textbf{91.8} &\textbf{83.4} &55.2 &13.8 &0.0 &\textbf{44.3} &\textbf{68.7} &\textbf{68.1}\\
    \hline
    \hline

\end{tabular}
}
\caption{Comparison with the state-of-the-art methods on J-HMDB Sentences. Our proposed model outperforms the state-of-the-art methods for nearly all metrics. Top-2 results are highlighted in bold.}
\label{Table2}
\vspace{-5pt}
\end{table*}

The mask generation module of the algorithm in this chapter adopts the structure of U-Net\cite{ronneberger2015u} considering its successful performance. Our method predicts sequential masks after selecting kernels based on the matching scores, which greatly reduce the number of input object queries, and avoids well-prepared annotations to learn instance segmentation at first. 

\noindent{\textbf{$\bullet$ Coarse Mask Generation and Loss}}

After the Transformer decoder, the query features $F_q$ is obtained, which is based on the observation of the overall content from both visual and linguistic aspects.
We use two layers of MLP to map $F_q$ 
to get a set of dynamic convolution kernels $\Omega \in \mathbb{R}^{N_q\times (T\times K)\times \frac{H }{16}\times \frac{W}{16}}$. We convolve $\Omega$ with the language-guided visual features $F_{L\rightarrow V}$ obtained by the Transformer decoder to obtain coarse-grained masks ${\rm Mask}_{q,LR}$:
\begin{equation}
{\rm Mask}_{q,LR} = \Omega \otimes F_{L\rightarrow V}
\label{equation3_2_5}
\end{equation}
where $LR$ represent low-resolution, ${\rm Mask}_{q,LR} \in \mathbb{R}^{N_q\times (T\times K)\times \frac{H}{16}\times \frac{H}{16}}$. For each coarse-grained mask in ${\rm Mask}_{q,LR}$, we calculate the pixel-level IoU with the down-sampling ground truth, select the one with highest pixel-level IoU, denote it as ${\rm Mask}_{LR}$. Meanwhile, denote the indicators of positive/negative regions as $\Delta = [\delta_1, \delta_2, ..., \delta_{N_q}]$, where
$\delta_j = 1$ when the mask ${\rm Mask}_{LR}^j$ has the highest pixel-level IoU with the down-sampling ground truth and otherwise 0. 
We utilize an additional fully-connected layer to obtain matching scores $S \in \mathbb{R}^{N_q}$, with each element indicate the matching scores of the object query with the referring object.
\begin{equation}
S = \sigma_s(W\cdot [F_q;F_{V\rightarrow L}])
\end{equation}
where $W$ is the learnable parameter, $\sigma_s$ is softmax activation, and [;] represents concatenate operation. 
For each resulting coarse-grained video mask, we perform down-sampling operation to the ground truths to supervise the generation of the coarse-grained mask by a cross-entropy loss. Meanwhile, we also incorporate the matching loss to get the total loss at low resolution:
\begin{equation}
\mathcal{L}_{LR} = \sum_{t=1, k=1}^{T, K} [\beta \cdot CE({\rm Mask}_{LR}^{t,k}, {\rm GT}_{LR}^{t,k})-\sum_{j=1}^{N_q} \delta_j^{t,k} \cdot {\rm log} S_j^{t,k}]
\label{equation3_2_4}
\end{equation}
where $i$ indicates $i$-th GoP, $j$ indicates $j$-th the P-frame, ${\rm GT}$ represents ground-truth, $CE$ represents cross-entropy loss, and $\beta$ is the hyper-parameter which aims to control the balance of two losses.

\noindent{\textbf{$\bullet$ Refined Mask Generation and Loss}}

After getting the coarse-grained ${\rm Mask}_{LR}$, we combine it with the visual output of the Transformer encoder $F_{L\rightarrow V} \in \mathbb{R}^{(T\times K)\times \frac{H}{16}\times \frac{W}{16}}$ and the feature encoder corresponding scale features first. Then, we leverage the up-sampling and convolution operation to get up-scale mask.

Next, two additional up-sampling and convolution operation is applied, each time we concatenate the mask with corresponding scale feature of the dual-path dual-attention output. After two additional operation, we get the fine-grained ${\rm Mask}_{HR}$, and we leverage the annotations and two losses to supervised the fine-grained mask generation. The formulation is:
\begin{equation}
\mathcal{L}_{HR} = \sum_{t=1, k=1}^{T, K} [CE({\rm Mask}_{HR}^{t,k}, {\rm GT}_{HR}^{t,k}) + Dice({\rm Mask}_{HR}^{t,k}, {\rm GT}_{HR}^{t,k})]
\label{equation3_2_4}
\end{equation}
where $HR$ indicates the high-resolution, $Dice$ is the dice loss.

Thus, the total loss in the proposed method is:
\begin{equation}
\mathcal{L} = \mathcal{L}_{HR} + \mathcal{L}_{LR}
\label{equation3_2_4}
\end{equation}
During inference, we select the highest matching score and its corresponding coarse-grained mask. Then we leverage the coarse-grained mask to get the final refined mask with the output of transformer encoder and feature encoder. 





\section{Experiments}

\begin{figure*}[t]
\centering
\includegraphics[width=0.9\linewidth]{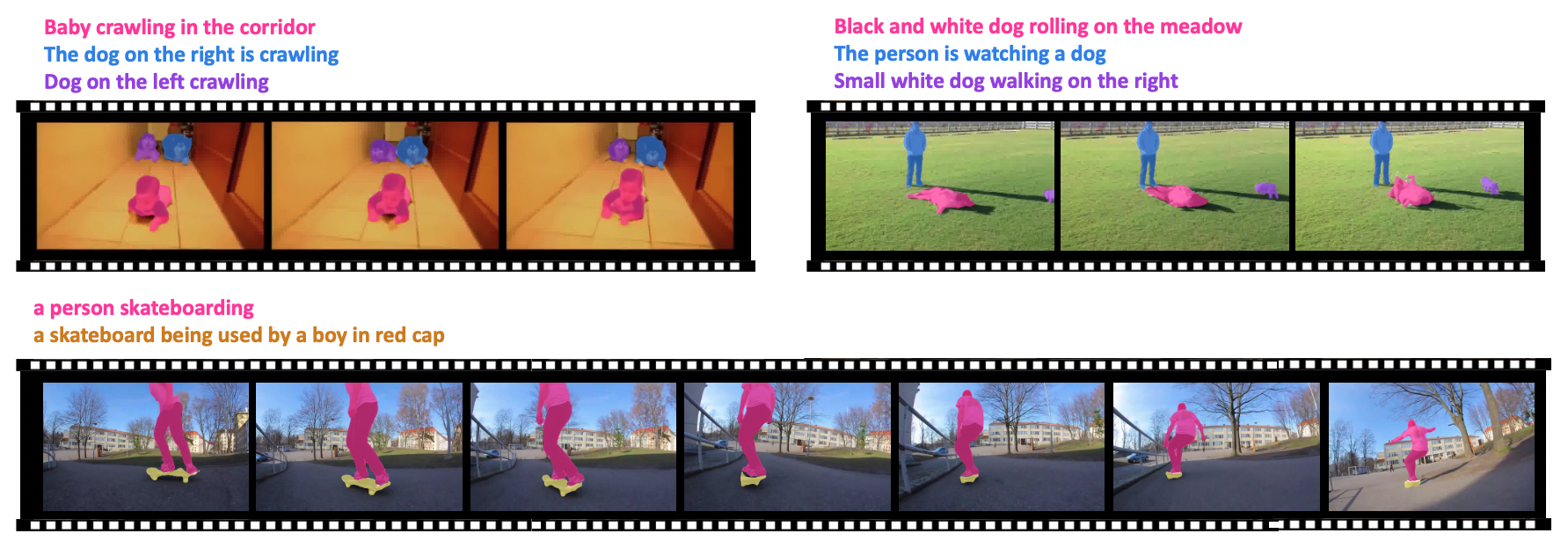}
\caption{Qualitative Results of our proposed method. The above two are the results on A2D Sentences; whereas, the bottom one is the results on Refer-YouTube-VOS. Best viewed in color.}
\label{fig4}
\end{figure*}

\subsection{Datasets and evaluation metrics}

We conduct extensive experiments on three referring video segmentation benchmarks including Actor-Action (A2D) Sentences, J-HMDB Sentences and Refer-YouTube-VOS.

{\textbf{A2D Sentences.}} The original A2D dataset contains 3,782 videos which is collected from YouTube. 
Following \cite{xu2015can}, we split the dataset into 3,036 training videos and 746 testing videos. 
Gavrilyuk \textit{et al.} \cite{gavrilyuk2018actor} augments the original A2D dataset to the A2D Sentences by providing a sentence description to each actor and its action in a video. There are totally 6,655 sentences. 
We train our network on the training split of A2D Sentences and test its performance on the testing split of A2D Sentences and J-HMDB Sentences for all experiments in this paper.

{\textbf{J-HMDB Sentences.}} The J-HMDB dataset contains 928 video clips of 21 different actions with mask annotations. Through Gavrilyuk \textit{et al.} \cite{gavrilyuk2018actor} extending, J-HMDB Sentences has 928 corresponding sentence descriptions for 928 videos. 

{\textbf{Refer-YouTube-VOS.}} To referring video object segmentation task, \cite{seo2020urvos} constructs a large-scale video object segmentation dataset, which is called Refer-Youtube-VOS, with descriptive sentences. The dataset has 4,519 high-resolution videos containing 94 common object categories. There are pixel-level instance segmentation annotations every 5 frames in each 30fps video, and their duration is about 3 to 6 seconds. 

{\textbf{Evaluation metrics.}} On A2D Sentences and J-HMDB Sentences datasets, following previous works \cite{gavrilyuk2018actor, wang2019asymmetric, wang2020context, mcintosh2020visual, ning2020polar, hui2021collaborative, ye2021referring, liu2021cross, chen2021cascade}, Overall IoU, Mean IoU, Precision@$K$ (P@$K$), and mAP are adopted as metrics to evaluate the performance. It is worth noting that all IoU in our task is the pixel-wise Intersection-over-Union. The precision@$K$ computes the percentage of samples whose IoU is higher than threshold $K$ ($K=\{0.5, 0.6, 0.7, 0.8, 0.9\}$). The mAP reports mean average precision (mAP) over 0.50:0.05:0.95. The overall IoU reports the ratio of the overall intersection area of all testing samples to the total union areas; while the mean IoU reports the average means of IoU over all the testing samples. On Refer-YouTube-VOS dataset, we leverage the standard evaluation metrics on this dataset, region similarity ($\mathcal{J}$), contour accuracy ($\mathcal{F}$) and their average value ($\mathcal{J} \& \mathcal{F}$). Since the annotations of validation set are not released publicly, we evaluate our method on the official challenge server.


\subsection{Implementation Details}\label{ImpleDets}

The video clip input for this work consists of 36 frames, including 3 I-frames, and each I-frame is followed by 11 the P-frames, including motion vectors and residuals. In this work, all frame inputs are resized to 320*320. The dimension of visual features extracted by Video-Swin-T is 20*20*384, the dimension of visual features extracted by ResNet-18 are 20*20*512, and the dimension of language features extracted by BERT is 20*768 (each language description is padding to 20 words). All experiments were implemented using PyTorch. Some other settings: $D$ is 512, learning rate is set to 0.0001, learning momentum is 0.9, weight decay is 0.0005, optimizer is SGD-momentum, training steps and epoch are 1000 and 60, respectively. $\beta$ is set to 0.1 in this work.


\subsection{Main Results}

{\textbf{Results on A2D Sentences.}} 
The results are shown in Table \ref{Table1}. From the resuls, we can see that our method is the only method that achieves the best result on both two major metrics, mean IoU and Overall IoU, and top-2 results on mAP, and also, we can observe that our method achieves the state-of-the-art results on almost all metrics, and achieves a substantial improvement. Thus, the effectiveness of our proposed method can be illustrated. 

To further illustrate our method's outperformance, we test the run time of our method and MTTR\cite{botach2021end}, which is method with the best performance so far and fast inference speed. On PC\footnote{Some details about our PC: 1) SSD: Samsung EVO 970 Plus; 2) CPU: AMD 3950X and 3) GPU: RTX 3090.}, our method achieves 77 fps\footnote{The inference speed is tested on the original codes without any quantification or accelerate operation.}; while, MTTR can only run on 52 fps. The speed of our method is about 48\% higher. 
Video decompression on our PC is very fast, running at over 1,000 fps, and it doesn't make a big difference to MTTR's inference speed. 
However, on the mobile terminal, video decompression requires a lot of time and computation. Here's some statistics, for our mobile terminal (Jetson Xaiver NX), decompressing 360p A2D video runs at 50 fps, and decompressing 1080p A2D video runs at only 8 fps. 
Especially in online segmentation on mobile terminal, video decompression hinders the running time of existing methods, and since our method is directly applied to compressed video, there is little impact on our method.


{\textbf{Results on J-HMDB Sentences.}} We evaluated the generalization ability of the proposed method
by directly testing the model, which is trained on A2D Sentences, on J-HMDB Sentences without any fine-tuning. The results are shown in Table \ref{Table2}, from the resuls, we can see that our method is the only method that achieves top-2 results in all three major metrics, mean IoU, Overall IoU and mAP.

{\textbf{Results on Refer-YouTube-VOS.}} Refer-YouTube-VOS is the most challenging dataset so far. The results on Refer-YouTube-VOS are shown in \ref{Table3}. 
From the results, it can be observed that our method achieves the best results on the overall $\mathcal{J}\& \mathcal{F}$ metric when compared to the existing works. 
On metric $\mathcal{J}$, our method is much better than the existing state-of-the-art methods, which we believe is because our method makes good use of multi-scale mask generation and can process more frames at a time when compared to the existing methods.

{\textbf{Qualitative Results.}} The qualitative results of our proposed method are shown in Fig. \ref{fig4}. The above two results are of A2D Sentences; whereas, the bottom results are of Refer-YouTube-VOS. It can be seen that our method segments right video objects in several challenging situations, i.e., complex video scenes, pose variation, occlusion and partially out of camera. Meanwhile, it can be seen that our segmentation results are of good quality and have smooth edges.

\subsection{Ablation Studies}

\begin{table}[t]
\centering
\resizebox{0.9\linewidth}{!}{
    \begin{tabular}{ccccc}
    \hline
    \hline
    Methods &Venue &$\mathcal{J}\& \mathcal{F}$ &$\mathcal{J}$ &$\mathcal{F}$\\
    \hline
    URVOS \cite{seo2020urvos}   &ECCV20   &47.23 &45.27 &49.19\\
    CMPC-V \cite{liu2021cross}  &T-PAMI21 &47.48 &45.64 &49.32\\
    YOFO \cite{li2022you}       &AAAI22   &48.59 &47.50 &49.68\\
    MTTR \cite{botach2021end}   &CVPR22   &55.32 &54.00 &\textbf{56.64}\\
    \hline
    Ours  &-  &\textbf{55.63} &\textbf{54.75} &56.51\\
    \hline
    \hline
\end{tabular}}
\caption{Comparison with the state-of-the-art methods on Refer-YouTube-VOS.}
\label{Table3}
\end{table}

\begin{table}[t]
\centering
\resizebox{0.95\linewidth}{!}{
\begin{tabular}{cccc}
\hline
\hline
Methods &mAP &Overall IoU &Mean IoU\\
\hline
Baseline+CoViAR \cite{wu2018compressed}   &40.2  &67.9  &57.3\\
Baseline+DMC-Net \cite{shou2019dmc}  &43.1  &69.4  &59.5\\
Ours  &\textbf{47.1} &\textbf{72.6} &\textbf{63.2}\\
\hline
\hline
\end{tabular}}
\caption{
The results of ablation studies on A2D Sentence to discuss on the effectiveness of dual-path dual-attention module.}
\label{Table4}
\end{table}

{\textbf{Discussion on the effectiveness of Dual-path Dual-attention module.}} 
We verify the effectiveness of our dual-path dual-attention module in this section. The details of every compared method are listed as follows:
{\textbf{1) Baseline+CoViAR \cite{wu2018compressed}}} exchanges our the encoder of features from compressed video and dual-path dual-attention into CoViAR method, which concat the I-frame feature and the P-frame feature. 
{\textbf{2) Baseline+DMC-Net \cite{shou2019dmc}}} is similar to CoViAR. The difference is that we use another resnet-18 to extract DMC features. From the results, we can see that our method achieves large advances to the existing encoder of features from compressed video. A possible reason is that our dual-path dual-attention focuses on enhancing the visual features in the spatial dimension, and at the same time, the encoder of the reference frames (I-frames) in this work adopts the latest and most powerful Video-Swin-T, which makes our video features good for segmentation. Meanwhile, the lightweight feature extractor makes our network segment 36 frames at a time, which helps maintain segmenting results coherent across the video to further improve the segmentation results.

{\textbf{Discussion on the effectiveness of query-based cross-modal Transformer.}} In this section, we evaluate the effectiveness of the query-based cross-modal Transformer by replacing it with the existing method's module. We leverage the very recent method MTTR\cite{botach2021end} to replace our Transformer and mask generation with their query-based segmentation and instance sequence matching module. The only difference between two comparing methods is that the number of video frames training at a time. The results are shown in Table \ref{Table5}. As illustrated, our method achieves better segmenting results than both two comparing methods. Since MTTR adopted Video-Swin-T to extract all video frames features, their features are stronger than ours. A possible reason for achieving better performance is that our features from compressed video are good for learning segmenting one object at a time since instance segmentation requires more powerful visual features.

\begin{table}[t]
\centering
\resizebox{0.95\linewidth}{!}{
\begin{tabular}{cccc}
\hline
\hline
Methods &mAP &Overall IoU &Mean IoU\\
\hline
Baseline+MTTR (w=8) \cite{botach2021end}   &42.2  &68.4  &59.3\\
Baseline+MTTR (w=10) \cite{botach2021end}  &43.7  &70.3  &61.1\\
\hline
Ours  &\textbf{47.1} &\textbf{72.6} &\textbf{63.2}\\
\hline
\hline
\end{tabular}}
\caption{The results of ablation studies on A2D Sentence to discuss on the effectiveness of query-based cross-modal Transformer.}
\label{Table5}
\end{table}

\begin{table}[t]
\centering
\begin{tabular}{cccc}
\hline
\hline
Methods &mAP &Overall IoU &Mean IoU\\
\hline
$N_q=1$  &44.7    &71.8  &60.1\\
$N_q=3$  &45.9    &\textbf{73.1}  &62.4\\
$N_q=5$  &\textbf{47.1}  &72.6 &\textbf{63.2}\\
$N_q=7$  &45.2    &71.3  &60.7\\
$N_q=9$  &45.5    &71.0  &61.6  \\
\hline
\hline
\end{tabular}
\caption{The results on A2D Sentence to investigate the influence of object queries' number.}
\label{Table6}
\end{table}

{\textbf{Effectiveness of the number of object queries.}} The results are shown in Table~\ref{Table6}. We can see that as the number of object queries increases from 1 to 5, the performance becomes better. All evaluation metrics reached the optimal value when $N_q=5$. When the number of queries continues to increase, the performance degrades and becomes to converge since the performance of $N_q=7$ and $N_q=9$ are quite the same. 
This is because our convolution kernel is generated and selected by a hard selection method. 
 When the number of queries increases, the selection matching accuracy will inevitably decrease, thus affecting the segmentation effect. 
 When the number of queries is small, the object query is not enough to cover the objects of the entire video scene, which affects the segmentation results.

\section{Conclusion}
In this paper, we focus on the new perspective of referring video object segmentation task, which aims to apply referring object segmentation to the compressed video domain. To address the problem, we propose a multi-attention network which consists of dual-path dual-attention module and a query-based cross-modal Transformer module. The dual-path dual-attention module aims to generate powerful video features based on all three data modalities; whereas, the proposed query-based cross-modal Transformer leverage a new scheme for referring video segmentation, 
which leverages language to select one target kernel and thus leads to only one segmentation mask. This scheme dramatically reduces the number of input object queries and completely removes the complicated mask-matching procedure of the existing method and, therefore, boosts the speed.

\begin{acks}
This work was supported in part by the Italy–China Collaboration Project Talent under Grant 2018YFE0118400; in part by the National Natural Science Foundation of China under Grant U21B2038, and 61836002, 61931008, 61872333, 61976069, 62022083, 61902092, in part by the Youth Innovation Promotion Association CAS; and in part by the Fundamental Research Funds for Central Universities.
\end{acks}
\bibliographystyle{ACM-Reference-Format}
\bibliography{sample-sigconf}










\end{document}